\documentclass[11pt]{article}

\usepackage[final]{acl}

\usepackage{times}
\usepackage{latexsym}

\usepackage{microtype}

\usepackage{inconsolata}

\usepackage{graphicx}

\usepackage{amsmath,amssymb}
\usepackage{xurl} 
\usepackage{subcaption} 
\usepackage{fontspec}
\usepackage{booktabs}
\usepackage{xeCJK}
\usepackage{placeins}

\newcommand{\cond}[1]{%
  \begingroup
  \ttfamily
  \let\US\_
  \def\_{\US\allowbreak}
  #1\hspace{0pt}
  \endgroup
}

\title{Route-Induced Density and Stability (RIDE): Controlled Intervention and Mechanism Analysis of Routing-Style Meta Prompts on LLM Internal States}

\author{
Dianxing Zhang , Gang Li , Sheng Li \\
Digital China AI Research Institute \\
\texttt{zhangdxh@digitalchina.com}
}

\begin{document}
\maketitle
\begin{abstract}
Routing is widely used to scale large language models, from Mixture-of-Experts
gating to multi-model/tool selection. A common belief is that routing to a task
``expert'' activates sparser internal computation and thus yields more certain
and stable outputs (the Sparsity--Certainty Hypothesis). We test this belief by
injecting routing-style meta prompts as a textual proxy for routing signals in
front of frozen instruction-tuned LLMs. We quantify (C1) internal density via
activation sparsity, (C2) domain-keyword attention, and (C3) output stability
via predictive entropy and semantic variation. On a RouterEval subset with three
instruction-tuned models (Qwen3-8B, Llama-3.1-8B-Instruct, and
Mistral-7B-Instruct-v0.2), meta prompts consistently densify early/middle-layer
representations rather than increasing sparsity; natural-language expert
instructions are often stronger than structured tags. Attention responses are
heterogeneous: Qwen/Llama reduce keyword attention, while Mistral reinforces it.
Finally, the densification--stability link is weak and appears only in Qwen,
with near-zero correlations in Llama and Mistral. We present RIDE as a
diagnostic probe for calibrating routing design and uncertainty estimation.
\end{abstract}

\section{Introduction}
As large language models (LLMs) are increasingly deployed in search assistants, coding assistants, and multi-tool systems, routing has become a foundational mechanism. In Mixture-of-Experts (MoE) architectures, a lightweight gating network activates only a small subset of expert subnetworks conditioned on the input, achieving parameter-level sparsification and computational savings. In multi-model routing systems, a policy model selects among multiple LLMs, toolchains, or APIs to balance performance, cost, and safety. As routing structures grow more complex, a new challenge emerges: we must understand not only whether routing decisions are ``correct,'' but also how routing signals shape the backbone model’s internal representations and reasoning patterns.
A compelling intuition in current practice and informal discussions is that routing effectively selects a narrower, more specialized ``expert pathway'', making internal computation sparser and more focused, ultimately producing more certain and stable outputs. We formalize this intuition as the Sparsity--Certainty Hypothesis: when a model receives a clear routing signal (e.g., ``this is a math task''), its internal representations propagate along more selectively gated sparse pathways, suppressing task-irrelevant activations and thereby reducing output entropy and increasing cross-sample consistency.
This hypothesis has been implicitly adopted in various settings—for instance, using internal signals such as activation sparsity or attention concentration as proxies for uncertainty estimation, early exit, or routing decisions. However, modern instruction-tuned LLMs have highly complex internal structures that strongly depend on training details, and whether this hypothesis holds has not been systematically tested.
Directly ``opening up'' real MoE gating mechanisms or complex policy routers is often impractical. Engineering systems are typically tightly encapsulated, making expert activation dynamics difficult to access; moreover, modifying routers to insert diagnostic signals may compromise stability and safety in production. We therefore take a pragmatic alternative: instead of intervening on the router itself, we inject routing-style meta prompts at the input level as a textual proxy for routing decisions.
Concretely, we prepend different types of prefixes to frozen instruction-tuned LLMs, including structured route tags (e.g., [RouteTag=math]), natural-language expert instructions (e.g.,  "You are a Math Expert."), as well as control conditions such as incorrect tags and placebo tags. For each instance, we keep the backbone parameters and random seeds identical across prefix conditions and vary only the prefix, thereby constructing paired, controlled intervention experiments. By comparing internal activation density, domain-keyword attention, and output stability across conditions, we can analyze how ``routing-style signals'' modulate internal states.
Building on this idea, we propose RIDE (Route-Induced Density and Stability), centered on three research questions:
\begin{itemize}
\item RQ1 (Density effects): Under a frozen backbone, how do different routing-style meta prompts (route tags vs. expert prompts) change the density/sparsity of internal activations? Which layer segments are most affected?
\item RQ2 (Attention strategies): Do routing-style meta prompts change how the model leverages domain information? After receiving explicit labels, do models ``offload'' reliance on domain keywords, or instead increase their focus on keywords?
\item RQ3 (Densification--stability relation): At the instance level, is there a stable association between changes in internal densification (C1) and changes in output stability (C3)? Is this association consistent across models, and can it serve as a general proxy for routing decisions?
\end{itemize}
Importantly, the ``routing signals'' analyzed in this work are not the internal states of real MoE gating modules or multi-model routers; they are routing-style meta instructions injected as textual prefixes. Our conclusions thus constitute causal-style interventional evidence from controlled experiments, rather than full-fledged structural causal identification.
\subsection{Contributions}
Within this framework, our main contributions are:
\begin{itemize}
\item \textbf{RIDE metrics and a unified controlled-intervention pipeline.}
 We design a suite of routing-style meta prompts and control conditions (control / correct tag / incorrect tag / placebo tag / expert instruction), construct paired interventions on three open-source instruction-tuned LLMs, and define three metric families—C1 (density), C2 (domain-keyword attention), and C3 (output stability)—as a systematic toolkit for analyzing routing-signal effects.
\item \textbf{Model heterogeneity in densification and attention responses to meta prompts.}
 Experiments show that all models exhibit significant densification responses to task-oriented meta prompts in the Early/Middle layers, while natural-language expert instructions induce stronger densification than structured route tags in most settings. At the attention level, different models can even exhibit opposite strategies: Llama/Qwen align more with cognitive offloading, whereas Mistral displays attention reinforcement.
\item \textbf{A systematic test of the ``densification $\Rightarrow$ stability'' chain with informative negative results.}
 We observe a small-to-moderate positive correlation consistent with densification--stability coupling in Qwen3-8B, but the link is weak or statistically insignificant at the instance level for Llama-3.1-8B-Instruct and Mistral-7B-Instruct-v0.2. These results suggest limited empirical support for the assumption that internal density serves as a cross-model, general-purpose uncertainty measure or routing signal. Consequently, RIDE is better positioned as a diagnostic probe for revealing heterogeneous model responses to routing-style meta prompts.
\end{itemize}

\section{Background and Related Work}
We briefly review three lines of work---routing and routing benchmarks, sparsity--certainty assumptions, and interpretability/meta-prompt analysis---and position RIDE among them.

\paragraph{Routing mechanisms and benchmarks.}
In Mixture-of-Experts (MoE) models, a gating network activates a small subset of experts for conditional computation, with extensive work on scaling and training stability (e.g., load balancing and regularization). \citep{Shazeer2017SparselyGatedMoE,Fedus2022SwitchTransformers,Du2022GLaM,Lepikhin2020GShard}
However, these advances are largely evaluated by downstream performance and FLOPs, and rarely analyze how gating signals shape intermediate representations. \citep{Zoph2022STMoE,Rajbhandari2022DeepSpeedMoE}
In multi-model/tool routing, a policy selects among models or tools under capability--cost trade-offs. \citep{Chen2023FrugalGPT,Schick2023Toolformer,Yao2023ReAct}
Recent benchmarks and learned routing strategies enable systematic evaluation (e.g., across tasks, difficulty, and cost), but mainly assess input--output selection quality rather than internal-state effects. \citep{Hu2024RouterBench,Huang2025RouterEval,Song2025IRTRouter}
In contrast, we do not build or optimize routers; we treat routing-style meta prompts as controllable textual proxies of routing signals and analyze their internal effects. \citep{Hendel2023TaskVectors,Liu2024InContextVectors,Stolfo2025ActivationSteering}

\paragraph{Sparsity and certainty.}
Across compression, MoE, and adaptive inference, activation sparsity/attention concentration and entropy are often used as proxies for specialization or uncertainty, and as signals for early exiting or model switching. \citep{Frankle2019LotteryTicket,Sanh2020MovementPruning,Fedus2022SwitchTransformers,Du2022GLaM,Zhou2020PABEE,Xin2020DeeBERT,Chen2023FrugalGPT,Vazhentsev2022UncertaintyTransformers}
These practices implicitly suggest that ``sparser/more concentrated internal states $\Rightarrow$ more stable outputs.'' \citep{Zhai2023AttentionEntropyCollapse}
We formalize this as the Sparsity--Certainty Hypothesis and test it via controlled meta-prompt interventions on frozen instruction-tuned LLMs, separating mechanism probing from performance optimization. \citep{Chung2024ScalingInstructionFinetunedLMs,Stolfo2025ActivationSteering}

\paragraph{Interpretability and meta-prompt analysis.}
Mechanistic interpretability probes internal computation via activations, attention, and circuit-level discovery/attribution. \citep{Meng2022LocatingEditingFacts,Conmy2023ACDC,Ameisen2025CircuitTracing,Zhang2025EAPGP,Jain2019AttentionNotExplanation,Wiegreffe2019AttentionNotNotExplanation}
Recent work also shows that prompts/instructions can steer internal representations and induce measurable distraction/attraction phenomena that can be localized to specific heads or circuit components. \citep{Hendel2023TaskVectors,Liu2024InContextVectors,Niu2025LlamaSeeLlamaDo}
We follow this direction but focus on routing-style meta prompts and, unlike single-model qualitative studies, provide a unified multi-model intervention pipeline to compare heterogeneity along density--attention--stability pathways. \citep{Huang2025RouterEval,Hu2024RouterBench,Rimsky2024CAA,Yu2025BackAttention,Meng2022LocatingEditingFacts}

\section{RIDE: Routing-Style Meta Prompts and Experimental Design}
\subsection{Routing-Style Meta Prompts}We inject routing-style meta prompts by prepending short prefixes to the user instruction. For each instance, we construct five prefix conditions:
1.\textbf{control}:
No routing-related prefix is added; the task instruction is fed directly.
2.\textbf{\texttt{tag\_correct}}:
 We add a structured tag that matches the instance domain, e.g., [RouteTag=math] or [RouteTag=code]. The tag format is fixed, with the internal field indicating the domain.
3.\textbf{\texttt{tag\_wrong}}:
 We add a tag that does not match the instance domain (e.g., [RouteTag=code] before a math problem) to probe intervention effects of ``incorrect routing signals.''
4.\textbf{\texttt{tag\_placebo}}:
 The tag has the same surface format as \texttt{tag\_correct}, but the internal string carries no recognizable semantics (e.g., random tokens), controlling for prefix length and formatting effects.
5.\textbf{\texttt{instr\_expert}}:
 We use natural-language expert instructions such as ``You are a Math Expert.'' or ``You are a coding assistant.'', representing widely used role-setting prompts.
 
All conditions share the same core instruction text, frozen backbone parameters, and decoding configuration; only the prefix differs. Thus, each instance forms a strictly paired intervention experiment across the five conditions, and differences can be interpreted as controlled intervention effects of the ``routing-style signal.''
\subsection{Dataset and Domain Partitioning}We construct our experimental samples from the RouterEval dataset and focus on three sub-domains:
\textbf{Math}: multi-step mathematical reasoning problems;
\textbf{Format / IFEval}: tasks with strict output-format constraints;
\textbf{Commonsense}: primarily multiple-choice commonsense reasoning tasks.
Following RouterEval's easy/hard difficulty split, we sample instances from the training split within each sub-domain, ensuring that every model sees exactly the same inputs under all prefix conditions. To avoid data leakage, we use only the official training partition and keep the per-domain sample size on the order of hundreds to a thousand. Appendix A reports detailed statistics of sample distributions by domain and difficulty.
\subsection{Models and Decoding Configuration} 

We evaluate three open-source instruction-tuned models:
\textbf{Llama-3.1-8B-Instruct},
\textbf{Mistral-7B-Instruct-v0.2},
\textbf{Qwen3-8B}.
The three models have similar parameter scales, are instruction-tuned, and perform well across multiple tasks. We use a unified decoding configuration (temperature, top-p, maximum length, etc.). For each instance and each fixed random seed, we generate K candidate outputs to estimate output entropy and semantic variation. All prefix conditions share the same set of random seeds, minimizing the influence of sampling randomness on paired differences.
\subsection{Representation Sampling and Layer-Segment Partitioning}We sample hidden states from every layer and uniformly partition layers by depth into three segments:
- Early: the first third of layers closest to the input embeddings;
- Middle: the middle third;
- Late: the final third closest to the output head.
Within each segment, we aggregate over the token dimension (e.g., by mean or max pooling) to obtain a representative hidden vector for computing metrics such as Hoyer sparsity and Top-k energy. For attention-based metrics, we primarily focus on the last layer (or several late layers) as an approximation of how much the decoder attends to domain keywords during decision making.

\section{Metrics and Analysis Methods}
We define three metric families: \textbf{C1} (activation sparsity/density),
\textbf{C2} (domain-keyword attention), and \textbf{C3} (output stability).
Unless otherwise noted, all metrics are computed per instance and per prefix
condition, then compared via paired differences.

\subsection{C1: Activation Sparsity / Density}
We quantify the sparsity of a hidden vector $\mathbf{h}\in\mathbb{R}^d$
using Hoyer sparsity:
$\mathrm{Hoyer}(\mathbf{h})=
\left(\sqrt{d}-\|\mathbf{h}\|_1/\|\mathbf{h}\|_2\right)/(\sqrt{d}-1)\in[0,1]$,
where larger values indicate higher sparsity (thus lower values indicate
\emph{densification}).
We aggregate token-level Hoyer scores over layer segments (Early/Middle/Late).
In addition, we compute the Top-$k$ energy ratio (TopK), i.e., the fraction of
$\ell_2$ energy captured by the top-$k$ dimensions after sorting by $|h_i|$.
We define \textbf{C1} as a segment-level \emph{combined metric family} based on
Hoyer and TopK; due to space, the main paper primarily reports Hoyer changes,
while full definitions and additional TopK results are provided in Appendix~B.

\subsection{C2: Domain-Keyword Attention}
To measure reliance on domain signals, we build a small keyword lexicon for
each sub-domain (e.g., math, code) and compute the \emph{attention share} paid
to matched keyword positions.
Concretely, from the last-layer attention matrix, for a query position $t$ we
sum the attention mass assigned to keyword token positions; we then average
over (relevant) positions and over instances to obtain the domain-keyword
attention share.
We report two complementary viewpoints:
\textbf{Prompt-last} (query at the last input token) and
\textbf{First-gen} (query at the first generated token), corresponding to
domain-information usage when ``finishing reading'' vs.\ ``starting to answer.''
Formal definitions are in Appendix~B.

\subsection{C3: Output Stability}
We characterize output stability with two complementary proxies.
\paragraph{Predictive entropy (Entropy).}
During decoding, we compute token-level softmax entropy and average over the
generated sequence; for stochastic decoding, we further average this quantity
over the $K$ generations to obtain a per-instance mean entropy.
Lower values indicate a more concentrated predictive distribution.
\paragraph{Semantic variation (SemVar).}
For $K$ stochastic generations from the same prompt, we compute pairwise
embedding similarities and use $\mathrm{Var}=1-\mathrm{mF1}$ as a variation
score; smaller values indicate higher semantic consistency across generations.
Implementation details (encoder choice, $K$, sampling settings) are deferred to
Appendix~C. In the main text, we analyze Entropy and SemVar trends separately.

\subsection{Paired Differences and Correlation Estimation}
Our primary analyses use instance-level paired differences.
For a metric $M$ and instance $i$, we define
$\Delta M_i = M_i(m) - M_i(\texttt{control})$ (e.g.,
$\Delta\mathrm{Hoyer}_i=\mathrm{Hoyer}_i(\texttt{tag\_correct})
-\mathrm{Hoyer}_i(\texttt{control})$).
Paired differences reduce variance from instance-specific difficulty.
We test $\Delta$ effects using paired $t$-tests or Wilcoxon signed-rank tests
(with Benjamini--Hochberg FDR correction for multiple comparisons).
Unless otherwise stated, correlations between $\Delta$ metrics are Pearson
correlations; Spearman results are reported in Appendix~D and are consistent
with the main conclusions.

\section{Experimental Results}
This section presents the experimental results and analyses for RQ1–RQ3 in turn.

\subsection{RQ1: How Do Routing meta prompts Change Internal Density?}

\begin{figure}[t]
  \centering
  \includegraphics[width=\columnwidth]{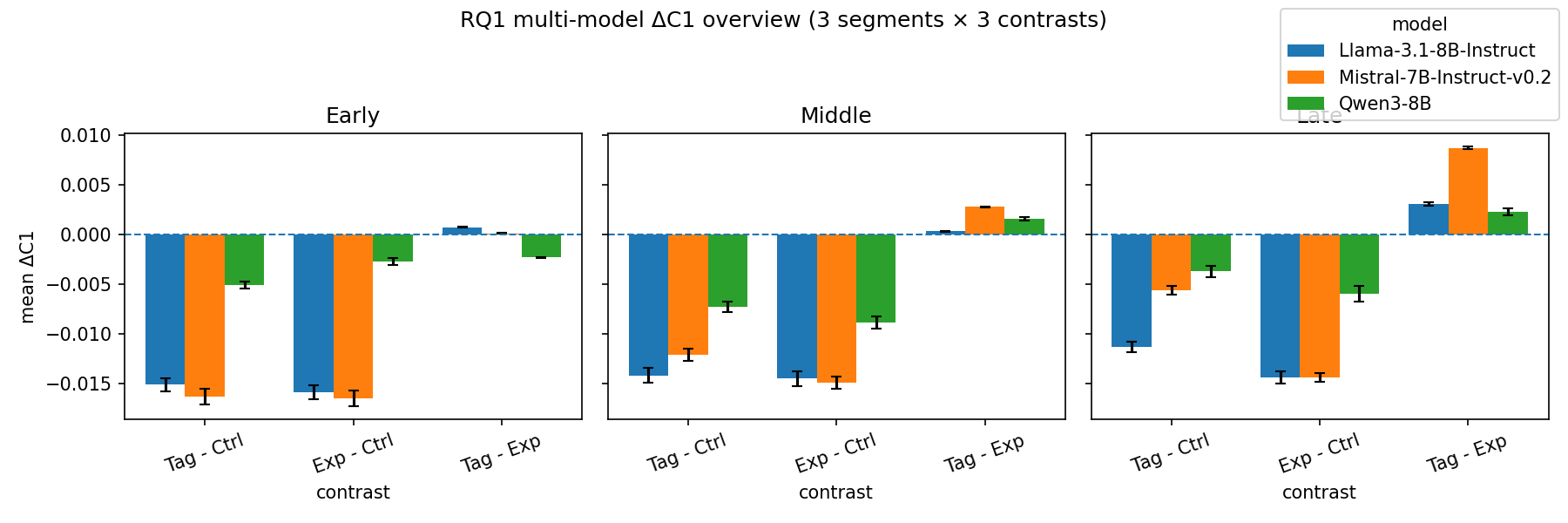}
  \caption{Multi-model overview of $\Delta \mathrm{Hoyer}$ across Early/Middle/Late segments.}
  \label{fig:rq1_hoyer_overview}
\end{figure}

Figure~\ref{fig:rq1_hoyer_overview} reports instance-level paired differences in Hoyer sparsity
($\Delta \mathrm{Hoyer}$) for \texttt{tag\_correct} and \texttt{instr\_expert} relative to \texttt{control},
as well as \texttt{tag\_correct} relative to \texttt{instr\_expert}.

\noindent\textbf{Densification vs.\ control.} Across all three models and most sub-domains, both \cond{instr\_expert} and \cond{tag\_correct}
yield negative $\Delta \mathrm{Hoyer}$ in the Early/Middle segments (typically $0.005$--$0.015$ in magnitude),
indicating a shift toward denser and more evenly distributed activations. In contrast, effects in the Late segment are markedly smaller or near zero, suggesting that routing-style
meta prompts primarily reshape earlier semantic representations rather than layers close to the output head.
As a sanity check, \texttt{tag\_placebo} shows $\Delta \mathrm{Hoyer}$ close to zero, implying that prefix length
or formatting alone cannot explain the densification effect (see Appendix~\ref{app:supp} for full statistics).

\noindent\textbf{Expert instructions are stronger than tags.}
Directly comparing \cond{tag\_correct} with \cond{instr\_expert}, we observe a consistently positive
global-average gap
$(\mathrm{Hoyer}(\texttt{tag\_correct}) - \mathrm{Hoyer}(\texttt{instr\_expert}) > 0)$
across all three models.
Since larger Hoyer values indicate higher sparsity, this means that natural-language expert instructions
induce stronger densification than structured route tags.
This reverses the naive expectation that short, formatted tags would behave as a closer proxy to ``true''
routing signals, and instead highlights the sensitivity of instruction-tuned LLMs to semantically rich
natural-language task framing (Appendix~B).

\subsection{RQ2: How Do Routing meta prompts Modulate Domain-Keyword Attention?}
\begin{figure*}[t]
  \centering
  \begin{subfigure}[t]{0.49\textwidth}
    \centering
    \includegraphics[width=\linewidth]{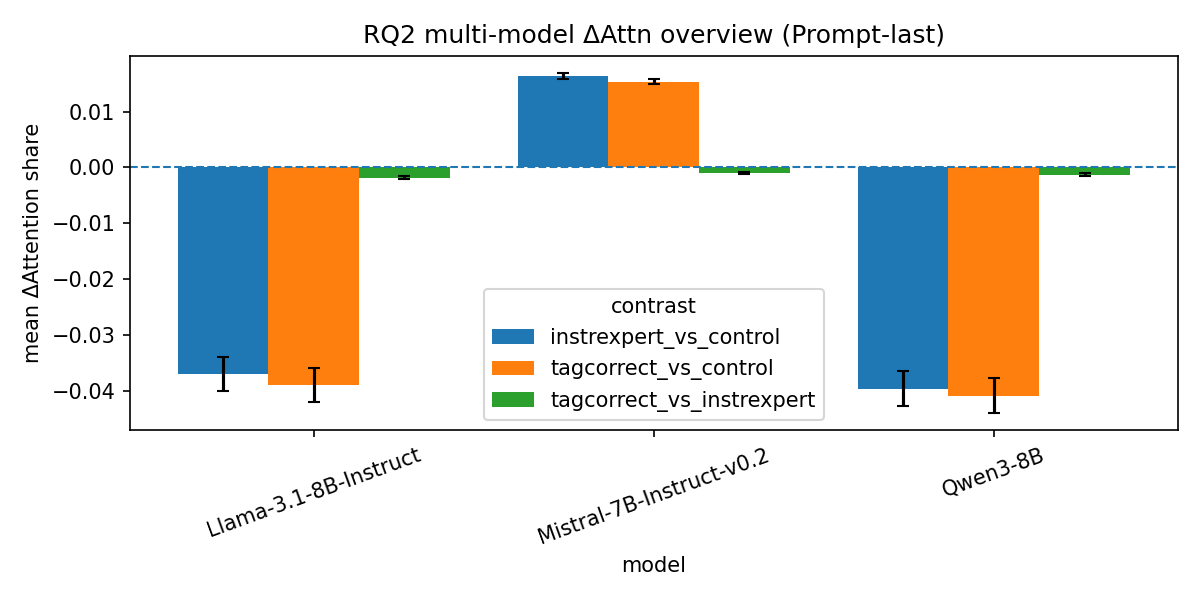}
    \caption{Prompt-last.}
    \label{fig:rq2_attn_a}
  \end{subfigure}\hfill
  \begin{subfigure}[t]{0.49\textwidth}
    \centering
    \includegraphics[width=\linewidth]{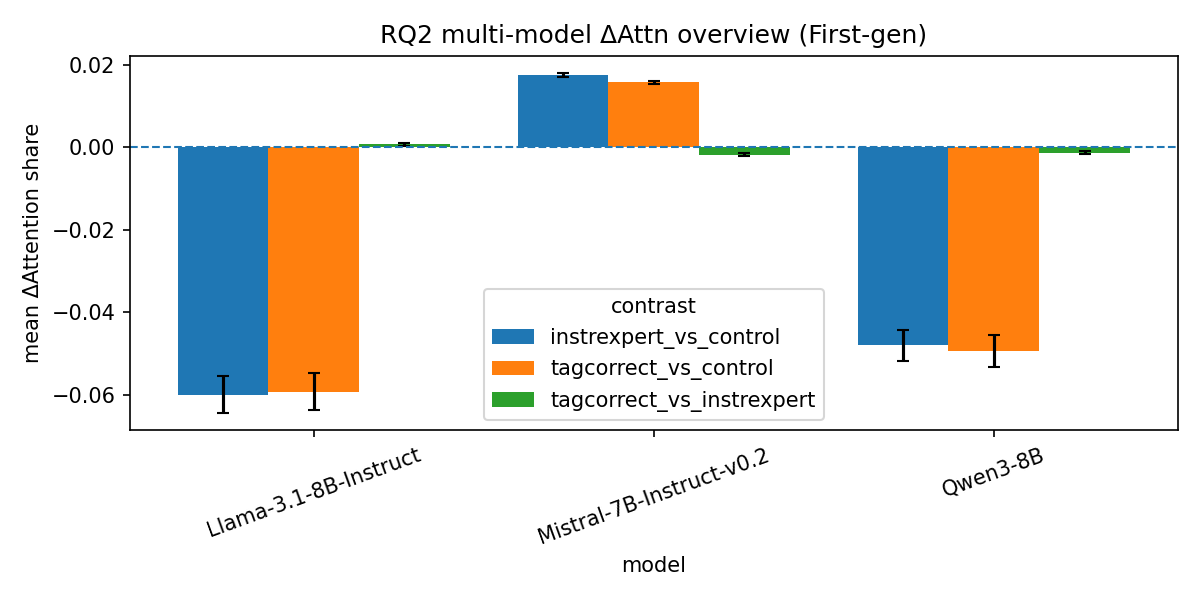}
    \caption{First-gen.}
    \label{fig:rq2_attn_b}
  \end{subfigure}
  \caption{Multi-model overview of $\Delta \mathrm{Attn}$ (mean $\pm$ SEM).
  Negative values indicate reduced attention share to domain keywords.}
  \label{fig:rq2_attn}
\end{figure*}

We measure C2 as the last-layer attention share assigned to domain-keyword tokens.
We estimate this share from two query viewpoints:
\textbf{Prompt-last} (query at the last input token) and \textbf{First-gen}
(query at the first generated token). Figure~\ref{fig:rq2_attn} reports
instance-level paired differences $\Delta \mathrm{Attn}$ under three contrasts
(\texttt{tag\_correct} vs.\ control, \texttt{instr\_expert} vs.\ control, and
\texttt{tag\_correct} vs.\ \texttt{instr\_expert}).

\paragraph{Model-specific attention strategies.}
Llama-3.1-8B-Instruct and Qwen3-8B show consistent \emph{decreases} in keyword
attention after adding routing-style prefixes (about $-0.04$ in Prompt-last and
$-0.05\sim-0.06$ in First-gen for Tag/Instr vs.\ control), suggesting an
\emph{offloading-like} pattern: the explicit routing cue partially substitutes
for lexical domain cues when the model begins responding. In contrast,
Mistral-7B-Instruct-v0.2 shows a stable \emph{increase} (about $+0.016$ in both
views), indicating \emph{reinforcement}: the model attends to the routing cue
\emph{and} the domain keywords more strongly rather than trading one for the other.
Across models, Tag and Instr typically have comparable magnitudes within each view.

\paragraph{\texorpdfstring{$\Delta \mathrm{Attn}$ vs.\ $\Delta \mathrm{C3}$}{Delta Attn vs. Delta C3}: weak coupling.}
Across contrasts and models, changes in keyword attention provide limited
explanatory power for output stability: correlations between $\Delta \mathrm{Attn}$
and $\Delta \mathrm{C3}$ are small overall (at best weak-to-moderate for
entropy in Qwen, and typically closer to zero for semantic variation; see Appendix~C.2).
Thus, keyword attention is better interpreted as a task-identification covariate
than as a direct driver of stability.

\subsection{RQ3: Linking Internal Densification to Output Stability}

RQ3 tests a core link in the Sparsity--Certainty Hypothesis: whether
instance-level changes in internal density ($\Delta \mathrm{C1}$) co-vary with
changes in output stability ($\Delta \mathrm{C3}$).
We instantiate $\mathrm{C1}$ with prompt-segment Hoyer sparsity and measure
$\mathrm{C3}$ by predictive entropy and semantic variation $(1-\mathrm{mF1})$.
For each model and each prefix contrast (\texttt{instr\_expert} vs.\ \texttt{control},
\texttt{tag\_correct} vs.\ \texttt{control}, and \texttt{tag\_correct} vs.\ \texttt{instr\_expert}),
we compute Pearson correlations on paired differences; full statistical details are in
Appendix~\ref{app:rq3_stats}.
Figure~\ref{fig:rq3_hoyer} summarizes the results.

\begin{figure*}[t]
  \centering
  \begin{subfigure}[t]{0.49\textwidth}
    \centering
    \includegraphics[width=\linewidth]{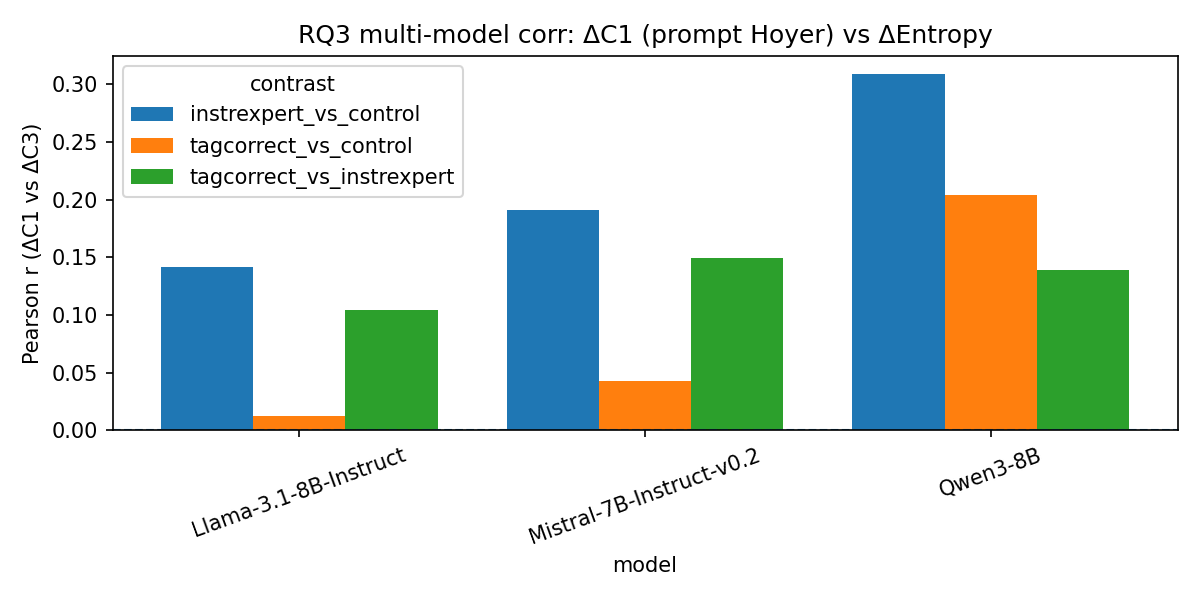}
    \caption{Entropy}
    \label{fig:rq3_entropy_a}
  \end{subfigure}\hfill
  \begin{subfigure}[t]{0.49\textwidth}
    \centering
    \includegraphics[width=\linewidth]{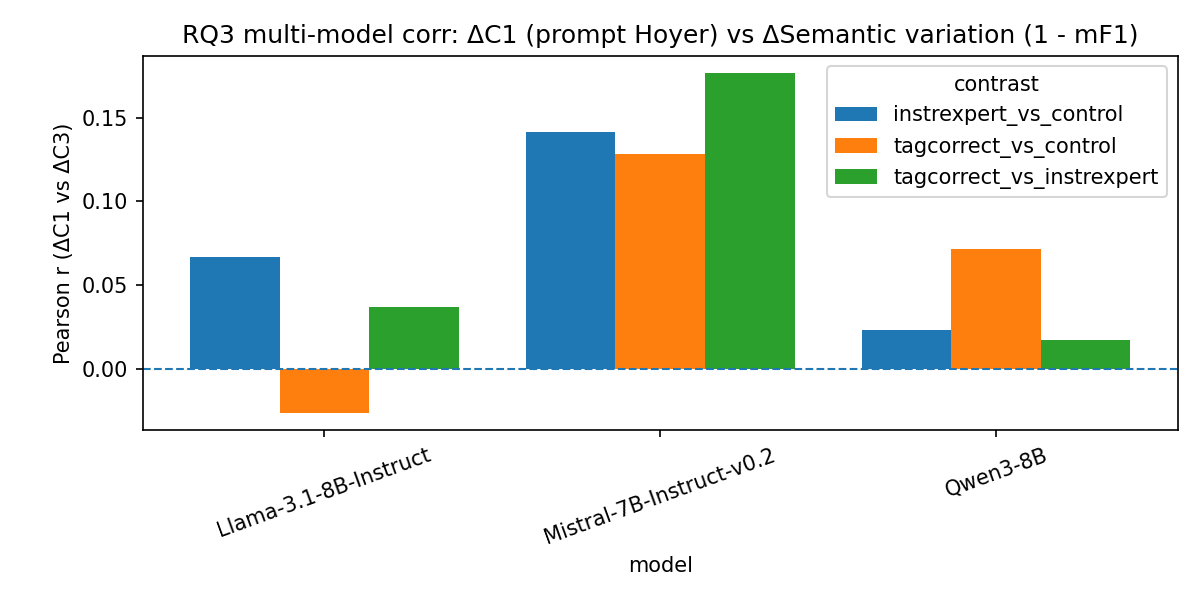}
    \caption{Semantic variation}
    \label{fig:rq3_variation_b}
  \end{subfigure}
  \caption{Instance-level correlations between $\Delta \mathrm{C1}$ (prompt Hoyer) and
  $\Delta \mathrm{C3}$ across models and contrasts.
  Positive values indicate that instances with larger decreases in Hoyer
  tend to also exhibit larger decreases in the corresponding instability measure
  (entropy or semantic variation).}
  \label{fig:rq3_hoyer}
\end{figure*}

\paragraph{Model-specific coupling.}
Qwen shows a consistent positive association between densification and lower entropy,
with correlations typically in the $0.2$--$0.3$ range across contrasts
(Appendix~\ref{app:rq3_stats}). In contrast, for Llama and Mistral the corresponding
correlations are near zero across contrasts, indicating that densification provides
little predictive signal for entropy changes in these models.

\paragraph{Semantic variation is weaker overall.}
Replacing entropy with semantic variation $(1-\mathrm{mF1})$ further attenuates the
relationship: Qwen retains only a small positive association, while Llama and Mistral
remain close to zero (Appendix~\ref{app:rq3_stats}).

\paragraph{Takeaway.}
The densification--stability link is not a model-invariant property: it appears in Qwen
but largely breaks in Llama and Mistral, suggesting that using internal density as a
general-purpose uncertainty proxy requires model-specific calibration.

\section{Discussion: RIDE as a Diagnostic Probe Rather Than a Universal Routing Law}
Taken together, RQ1–RQ3 paint a picture that is far more complex than the linear assumption "routing $\Rightarrow$ sparsity $\Rightarrow$ stability." Here we discuss the positioning and implications of RIDE from three perspectives.
\subsection{From the Sparsity--Certainty Hypothesis to a Model-Specific Density--Stability spectra}
Our original motivation was to test the intuition that ``effective routing selects sparser and more certain pathways.'' However, the experimental results show that:
\begin{itemize} 
\item At the density level: task-oriented meta prompts generally induce densification rather than increased sparsity;
\item At the stability level: a local ``densification--stability coupling'' is observed only in Qwen, while it is largely absent in Llama and Mistral.
\end{itemize}
Rather than a universal ``sparsity--certainty law,'' our findings suggest that different models form their own distinctive density--stability spectra under routing-style meta prompts: in some models, densification has a partially positive association with stability, whereas in others the two factors are nearly decoupled. This provides an important caution against directly transferring internal proxies across models.

\subsection{Implications for Routing Design and Uncertainty Estimation}
From an engineering standpoint, our results suggest that:
\begin{itemize} 
\item Routing signals should be calibrated in a model-specific manner.
 A proxy that works well for one model (e.g., C1) may completely fail for another, or even yield an opposite signal.
\item Natural-language instructions can themselves serve as strong routing signals.
 For instruction-tuned models without specialized fine-tuning, short expert instructions can substantially reshape internal density and attention structure, opening up design space for ``prompt-level routing.''
\item Incorrect or ambiguous routing tags may introduce additional instability.
 In the absence of a real router, indiscriminate use of ``expert tags'' may inject noise into internal states, potentially harming stability and safety.
\end{itemize}
In practical systems, RIDE can be used as a diagnostic tool: prior to deployment, one can run RIDE analyses on candidate models to determine which internal proxies carry stable semantics for that model, and only then consider using them for routing or uncertainty estimation.

\subsection{Scope and Limitations of RIDE}
RIDE offers a statistical diagnostic perspective that combines meta-prompt interventions with internal metrics, rather than precise identification of causal structure. We do not advocate directly using C1–C3 as decision signals in production systems. Instead, we suggest treating them as auxiliary tools for model understanding and iterative design—for example, selecting models that are better suited for ``prompt-level routing,'' or identifying architectures that are particularly sensitive to incorrect routing tags.

\section{Future Directions}
Future work can extend RIDE along several directions:
\begin{itemize}
\item Extend the framework to real MoE gating and multi-model routing logs, combining real routing signals with meta prompts;
\item Incorporate additional internal metrics (e.g., concept activations, specific neurons/subspaces) to enrich the characterization of density--stability dynamics;
\item Systematically evaluate the fairness and robustness of RIDE metrics on multilingual, multi-group, and safety-critical tasks;
\item Explore using RIDE as part of the routing-system design workflow, e.g., tracking how the density--stability spectra evolves before and after training.
\end{itemize}

\section{Conclusion}
We propose RIDE (Route-Induced Density and Stability), a framework that performs controlled interventions on frozen instruction-tuned LLMs via routing-style meta prompts, and systematically analyzes how routing signals affect internal density (C1), domain-keyword attention (C2), and output stability (C3). Based on large-scale experiments on a RouterEval subset across three open-source instruction-tuned models, we find that:
\begin{itemize}
\item Task-oriented meta prompts consistently induce densification of internal representations rather than increased sparsity;
\item Models exhibit pronounced heterogeneity in attention redistribution (cognitive offloading vs. attention reinforcement);
\item The ``densification $\Rightarrow$ stability'' link receives only limited support in a subset of models and lacks cross-model robustness.
\end{itemize}
Taken together, these findings suggest that directly treating internal density or attention concentration as a universal ``good routing signal'' or a general-purpose uncertainty proxy can be risky. We argue that RIDE is better positioned as a diagnostic probe: through controlled, statistical analyses, it reveals model-specific density--stability spectra under routing-style meta prompts, providing fine-grained evidence for routing design, uncertainty estimation, and model selection.

\section{Limitations}
This work has several important limitations (which we state explicitly to clarify the scope of generalization):
\begin{enumerate}
\item Proxy nature of routing signals and external validity.
 We use routing-style meta prompts as a textual proxy for real MoE gating or multi-model router signals. This proxy cannot capture routing distributions, logging features, or training couplings present in real systems; therefore, our conclusions primarily apply to intervention analyses of ``prompt-level routing signals.''
\item Limited coverage of tasks and models.
 Our experiments focus on a subset of RouterEval and three mid-sized open-source instruction-tuned models. Larger-scale models, different alignment strategies or specialist models, and different task distributions—especially long-context and tool-use settings—may exhibit different RIDE spectra.
\item Dependence on metric construction and decoding settings.
 The concrete implementations of C1–C3 (e.g., the exact Hoyer sparsity formulation, keyword-list coverage, sampling counts, and temperature) affect numerical scales. We provide several robustness checks in the appendix, but cannot exhaustively cover all configurations; using RIDE as a system signal requires model- and scenario-specific recalibration.
\item Boundaries of causal interpretation.
 Although we use paired interventions and control decoding and randomness, RIDE remains a ``causal-style'' statistical comparison rather than structured causal identification. We cannot fully rule out the influence of unobserved confounders (e.g., training history or alignment preferences) on the observed effects.
\end{enumerate}

\section*{Ethical considerations}
All experiments in this work are conducted using open-source models and publicly available datasets, without involving user privacy or sensitive data. The code and derived artifacts we release will not contain any personally identifiable information. RIDE is primarily intended as an analysis and diagnostic tool; any application that uses internal proxies for safety-critical decisions (e.g., filtering sensitive content, or routing decisions in medical or legal settings) should undergo rigorous safety evaluation and ethical review.

\bibliography{custom}

\appendix
\label{sec:appendix}

\section{Experimental and Implementation Details}
\subsection{Backbone Models and Hyperparameter Settings}
We conduct experiments on three open-source, instruction-tuned decoder-only LLMs. Throughout the appendix, we refer to them as Model-A / Model-B / Model-C, corresponding in the main text to:
\begin{itemize}
\item Model-A: Llama-3.1-8B-Instruct
\item Model-B: Mistral-7B-Instruct-v0.2
\item Model-C: Qwen3-8B
\end{itemize}
All three models follow an autoregressive Transformer architecture, stacking multi-head self-attention and feed-forward networks. They are pretrained primarily on English corpora and further instruction-tuned on multi-task instruction data. Unless stated otherwise, we adopt the following common settings:
\begin{itemize}
\item Frozen parameters: In all experiments, the backbone model parameters are fully frozen; interventions are applied only by modifying the input prefix (routing-style meta prompts).
\item Chat formatting: We use each model’s official tokenizer and chat/instruction formatting (system/user structure in chat/instruct mode), keeping the default system/assistant role templates unchanged.
\item Unified decoding hyperparameters: For a given model, decoding hyperparameters are kept identical across all prefix conditions.
\item Randomness control: For the same input instance, all prefix conditions share the same set of random seeds, ensuring that observed differences are primarily attributable to the prefix itself.
\end{itemize}

\subsection{RouterEval Preprocessing and Sampling}
\subsubsection{Constructing the RouterEval Training Table}
We first build a unified training/analysis table based on the public RouterEval dataset. Using a custom script, we generate a training-level sample table from the original structured RouterEval data. The command is as follows:
\begin{verbatim}
python build_routereval_dataset.py \
  --router_dataset_dir <dir_root> \
  --scenarios gsm8k,math,ifeval,\
  winogrande,mmlu \
  --difficulties easy,hard \
  --num_candidates 3,5,10 \
  --candidate_groups all_strong,\
  all_weak,strong_to_weak \
  --splits train \
  --output_csv <routereval_train_csv>
\end{verbatim}
\begin{itemize}
\item scenarios cover representative settings including arithmetic reasoning (gsm8k), mathematics/logic (math/mmlu), format compliance (ifeval), and commonsense reasoning (winogrande/mmlu), among others;
\item difficulties include easy and hard, spanning different difficulty levels;
\item The candidate configuration (\texttt{num\_candidates}, \texttt{candidate\_groups})  follows the original RouterEval settings to facilitate alignment with prior routing research.
\end{itemize}

\subsubsection{Building the Domain-Specific Items Table}
On top of this, we construct an items table for RIDE analysis. The command is:
\begin{verbatim}
python make_items_from_routereval.py \
  --input_csv <routereval_train_csv> \
  --output_csv <items_routereval_csv> \
  --domains math,ifeval,commonsense \
  --max_per_domain 200 \
  --drop_duplicates
\end{verbatim}
\begin{itemize}
\item domains selects three coarse-grained domains: Math/Code, Format/Verification (IFEval), and Commonsense;
\item max\_per\_domain caps the number of samples per domain at approximately 200, balancing diversity and computational cost;
\item drop\_duplicates removes duplicate or near-duplicate items.
\end{itemize}
The resulting <items\_routereval\_csv> serves as a unified ``analysis corpus table'' across models and prefix conditions in subsequent experiments, recording metadata such as the task text, domain labels, and source scenarios.

\subsubsection{Sample Filtering and Summary Statistics}
Based on the items table above, we apply lightweight filtering:
\begin{itemize}
\item We remove overly long instances (exceeding a predefined maximum token length) to avoid model-dependent truncation effects;
\item We remove extremely short instances (fewer than a minimum number of tokens) to prevent unstable estimates for metrics such as C2/C3.
\end{itemize}
After filtering, we retain approximately 200 instances per domain, forming a fixed analysis corpus. Table A.2 reports summary statistics for each domain, including the number of instances, the average token length, and the keyword matching rate.
 Because our focus is on internal change patterns under prefix interventions rather than generalization performance on this dataset, we do not further split this corpus into train/dev/test. Instead, we treat it as a fixed analysis set and compare different models on the same collection.

\subsection{Router-Style Meta-Prompt Templates and Tag Construction}
\subsubsection{Route Tags}
We adopt a unified structured format for route tags, as follows:
\texttt{[RouteTag=DOMAIN]}
where $\texttt{DOMAIN} \in \{\texttt{math}, \texttt{code}, \texttt{format}, \texttt{commonsense}\}$.
\begin{itemize}
\item tag\_correct: uses a tag that matches the instance domain (e.g., using [RouteTag=math] for a math problem);
\item tag\_wrong: uniformly samples an incorrect tag from the other domains (e.g., using [RouteTag=code] for a math instance).
\end{itemize}

\subsubsection{Placebo Tags}
Placebo tags are used to control for form factors—i.e., adding extra tokens in a similar format—without introducing meaningful semantics. We construct tags of the form:
\texttt{[RouteTag=XXXXX]}
where XXXXX is a meaningless string that rarely appears in the RouterEval corpus and has a length comparable to real route tags. Semantically, tag\_placebo should be approximately equivalent to control, differing only in surface form by the additional prefix tokens.

\subsubsection{Expert Prompts}
We design a small set of natural-language expert prompt templates for different domains, for example:
\begin{itemize}
\item Math
You are a Math Expert. Please solve the problem step by step.
 (A Chinese variant is also used: ``你是一名数学专家，请逐步推理并给出答案。'')
\item Format / IFEval
You are a formatting assistant. Please strictly follow the required output format.
\item Commonsense
You are a reasoning assistant. Please choose the most plausible option and briefly explain why.
\end{itemize}
In the main experiments, we use a fixed template per domain to reduce additional variance. In Appendix C, we conduct sensitivity checks with partially randomized template sets.

\subsubsection{Prefix Injection Location and Dialogue Structure}
Unless otherwise specified, all route tags and expert prompts are injected before the user prompt, e.g.,
[RouteTag=math]
<user prompt ...>
or
You are a Math Expert. Please solve the problem step by step.
<user prompt ...>
Other parts of the system/assistant roles (e.g., the system prompt and assistant prefix) are kept unchanged under each model’s default settings. Across prefix conditions, the only differences lie in the routing-style meta-prompt segment described above.

\subsection{Domain Keyword Lexicon (\texttt{DOMAIN\_KEYWORDS})}
The domain keyword lexicon is used to construct the C2 metric (domain-keyword attention share). We build it as follows:
\begin{enumerate}
\item  Collecting seed keywords
 For each sub-domain, we manually curate a small set of seed keywords:
\begin{itemize}
\item Math: equation, integer, sum, proof, factor
\item Code: function, variable, class, compile, error
\item Format: json, table, markdown, field, column
\item Commonsense: choose, option, likely, most, best).
\end{itemize}
 These seeds aim to cover the most central semantic concepts or task-triggering terms in the domain.
\item Automatic expansion and filtering:
\begin{itemize}
\item We use an auxiliary script to automatically expand the candidate list on the RouterEval training table using statistics such as TF-IDF / PMI;
\item We remove stopwords, punctuation, and extremely frequent but domain-agnostic words (e.g., the, is, do), and normalize case and inflectional forms.
\end{itemize}
\item Manual screening and finalization:
\begin{itemize}
\item  We perform a quick manual inspection of the expanded candidates and discard clearly inappropriate entries;
\item  We then form the final DOMAIN\_KEYWORDS dictionary and save it as a JSON file for downstream analysis scripts.
\end{itemize}
\end{enumerate}
For each instance, after tokenization, we match terms from DOMAIN\_KEYWORDS and record their token positions:
\begin{itemize}
\item When computing the domain-keyword attention share, the target set consists of these token positions;
\item If an instance has no keyword match under its domain, we define its domain-keyword attention share as 0 (since no attention mass falls on the keyword set), and mark it as ``missing keywords'' in the summary table (used to analyze the signal-to-noise ratio of C2).
\end{itemize}

\subsection{Extracting Activations and Attention}
\subsubsection{ Layer-Segment Partitioning}
Suppose the model contains L Transformer blocks. We evenly partition them into three segments:
\begin{itemize}
\item Early: Layers $1$ to $\lfloor L/3 \rfloor$;
\item Middle: Layers $\lfloor L/3 \rfloor + 1$ to $\lfloor 2L/3 \rfloor$;
\item Late: remaining layers.
\end{itemize}
For each segment, we aggregate across instances and tokens (e.g., by taking the mean) to obtain segment-level metrics.

\subsubsection{Caching Hidden States and Attention Weights}
For each instance and each prefix condition, we cache the following during the forward pass:
\begin{itemize}
\item The hidden vector at every token for every layer, $\mathbf{h}^{(\ell)}_{t}$;
\item Attention weights used for C2:
\begin{itemize}
\item[$\bullet$] Prompt-last: the attention over all prompt tokens at the final step before generation begins;
\item[$\bullet$] First-gen: the attention from the first generated token to all prompt tokens.
\end{itemize}
\end{itemize}
In practice, we enable output\_hidden\_states=True and output\_attentions=True in the inference API, or use framework-provided hooks to write intermediate results to files (e.g., Parquet / HDF5) for offline analysis.

\subsubsection{Attention Normalization and Keyword-Share Computation}
When computing the keyword attention share, we preprocess the attention vector of each layer as follows:
\begin{itemize}
\item We retain attention mass only over visible tokens, including system/user/prompt tokens, while excluding padding and special tokens (e.g., BOS/EOS);
\item We re-normalize over the visible-token set so that the weights sum to 1;
\item We sum the weights over positions that match DOMAIN\_KEYWORDS, yielding the layer-level keyword attention share;
\item We then aggregate across layer segments and instances to obtain the segment-level C2 metric.
\end{itemize}

\subsection{Decoding and Output Sampling (C3 Metric)}
\subsubsection{Decoding Hyperparameters}
For experiments related to output stability, we use stochastic sampling with temperature. Unless otherwise specified, we adopt the following unified hyperparameters:
\begin{itemize}
\item Sampling temperature T=0.7;
\item top-$p$ = 0.9;
\item No additional repetition penalty;
\item A fixed generation length limit max\_new\_tokens (e.g., 64).
\end{itemize}
 These settings are applied uniformly across all models and all prefix conditions.
 
\subsubsection{Multi-Sample Generation (Multi-sample Forward)}
To estimate C3 (predictive entropy and semantic variation), we generate K independent samples for each prompt under each prefix condition. In the main experiments:
\begin{itemize}
\item We use K=5 samples by default to estimate predictive entropy and semantic variation;
\item For the same instance, we reuse the same sequence of random seeds across different prefix conditions, ensuring that C3 differences are primarily attributable to prefix differences rather than sampling randomness.
\end{itemize}
Predictive entropy is computed at the token level and then averaged over the sequence. Semantic variation is computed based on similarities among sentence embeddings obtained from multiple samples; the formal definition is provided in Appendix B. Appendix C reports sensitivity analyses under different K values and temperatures.

\subsubsection{Sharding and Parallel Inference}
To improve computational efficiency, we shard <items\_routereval\_csv> by splitting the instances into multiple shards and run inference and metric computation in parallel on a multi-GPU setup. The overall workflow is:
\begin{itemize}
\item Use an auxiliary script to split the items CSV into multiple subfiles by row count (e.g., items\_part0.csv, items\_part1.csv, …);
\item Assign one or more shards to each GPU and run a unified inference script (which loads the model, applies prefixes, extracts hidden states and attention, and generates multi-sample outputs);
\item Save intermediate results for each shard (e.g., MFV metrics and generated texts) as structured files, and merge them during post-processing.
\end{itemize}
All shards are executed under the same configuration, ensuring comparability across GPUs and shards.

\subsection{Statistical Tests and Correlation Estimation}
To obtain causal-style paired evidence and to avoid over-interpreting any single scalar result, we adopt the following statistical procedures:
Paired difference tests
\begin{itemize}
\item For paired metric differences (e.g., $\Delta \mathrm{Hoyer}$, $\Delta \mathrm{Entropy}$, $\Delta \mathrm{KeywordShare}$), we use a paired t-test or the Wilcoxon signed-rank test to assess whether the differences significantly deviate from zero;
\item When appropriate, we report effect sizes (e.g., Cohen’s d) to reflect the practical magnitude of the differences.
\end{itemize}
Correlation analysis
\begin{itemize}
\item To quantify the linear relationship between $\Delta \mathrm{C1}$ and $\Delta \mathrm{C3}$, we use the Pearson correlation coefficient;
\item We estimate the 95\% confidence interval of Pearson r via the Fisher transformation;
\item In Appendix B, we additionally report Spearman’s rank correlation as a robustness check.
\end{itemize}
Multiple-comparison correction
\begin{itemize}
\item For the large number of tests across models, domains, and layer segments, we control the false discovery rate (FDR) using the Benjamini--Hochberg procedure, reporting significance at a unified q-level;
\item In the main paper, we report only conclusions with clear direction and magnitude that remain stable under multiple-comparison correction, avoiding over-interpretation of marginally significant results.
\end{itemize}

\section{Full Mathematical Definitions of Metrics }
This appendix provides the formal mathematical definitions of C1/C2/C3 and several auxiliary metrics used in Section 4 of the main paper. Notation is kept consistent with the main text.
\subsection{Notation}
\begin{itemize}
\item We denote an instance by $x$, and a routing prefix condition by $m \in \mathcal{M}$ (e.g., \texttt{control}, \texttt{tag\_correct}, \texttt{tag\_wrong}, \texttt{tag\_placebo}, \texttt{instr\_expert}).
\item The model has $L$ Transformer blocks. The hidden vector at layer $\ell$ and token position $t$ is $\mathbf{h}^{(\ell)}_{x,t}\in\mathbb{R}^d$.
\item Let the generated sequence length be $T_x$. The predictive distribution for the $t$-th generated token under condition $m$ is $p^{(m)}_t(v)$.
\item For attention, we denote the last layer by $\ell = L$, and its attention vector by
        $\mathbf{a}^{(L,m)}_{x,q}$, where $q$ is the query position (e.g., \texttt{prompt-last} or \texttt{first-gen}).
\item The set of token positions in the prompt segment is $\mathcal{P}_x$, and the set of domain-keyword positions is $\mathcal{K}_x \subseteq \mathcal{P}_x$.
\end{itemize}
Unless otherwise stated, instance-level metrics are first averaged over tokens and/or layers, and then subjected to statistical analysis over the set of instances.

\subsection{ C1: Activation Sparsity / Density}
In Section 4.1 of the main paper, C1 is described as a combined metric of segment-level Hoyer sparsity and Top-k energy. Here we provide precise definitions of both components and explain how C1 is constructed.
\subsubsection{Hoyer Sparsity}
Given a hidden vector $\mathbf{h}\in\mathbb{R}^d$, the Hoyer sparsity is defined as:
\[
\mathrm{Hoyer}(\mathbf{h})
= \frac{\sqrt{d}-\frac{\|\mathbf{h}\|_1}{\|\mathbf{h}\|_2}}{\sqrt{d}-1}.
\]
where $\mathbf{h}\in\mathbb{R}^d$ and $\mathrm{Hoyer}(\mathbf{h})\in[0,1]$.
Larger values indicate higher sparsity and stronger concentration on a small subset of dimensions.
For an instance x, prefix condition m, layer $\ell$, and token position t, the token-level Hoyer score is:
\[
s^{(\ell,m)}_{x,t} = \mathrm{Hoyer}\!\left(\mathbf{h}^{(\ell,m)}_{x,t}\right).
\]
We average over a layer segment. For example, for the Early segment (layer set $\mathcal{L}_{\mathrm{Early}}$), we define:
\[
\begin{aligned}
\mathrm{C1\_Hoyer}^{\mathrm{Early}}(x,m)
&= \frac{1}{\left|\mathcal{L}_{\mathrm{Early}}\right|\cdot\left|\mathcal{P}_x\right|} \\
&\quad \times
\sum_{\ell\in\mathcal{L}_{\mathrm{Early}}}
\sum_{t\in\mathcal{P}_x}
s^{(\ell,m)}_{x,t}.
\end{aligned}
\]
The Middle and Late segments are defined analogously. For the first-generated-token variant (C1-firstgen), we replace $\mathcal{P}_x$ with the token position corresponding to the first generated token.
The ``Hoyer part of C1'' in the main paper refers to this type of segment-level aggregation.

\subsubsection{Top-$k$ Energy Ratio}
Let $|h_{(1)}|\ge \cdots \ge |h_{(d)}|$ denote the components of $\mathbf{h}\in\mathbb{R}^d$
sorted by absolute value. The Top-$k$ energy ratio is defined as:
\[
\mathrm{TopK}(\mathbf{h})
= \frac{\sum_{i=1}^{k} h_{(i)}^2}{\sum_{i=1}^{d} h_i^2},
\qquad
k = \lfloor \alpha d \rfloor,
\]
with $\alpha=0.1$ by default.

In our setting, decreases in Hoyer (densification) are typically accompanied by decreases
in the Top-$k$ energy ratio, indicating that energy is more evenly distributed across a
larger number of dimensions.

Similarly, we define the token-level $\mathrm{TopK}^{(\ell,m)}_{x,t}$ and average over
layer segments and prompt-token positions. For example, for the Early segment:
\[
\begin{aligned}
\mathrm{C1\_TopK}^{\mathrm{Early}}(x,m)
&= \Bigl(\left|\mathcal{L}_{\mathrm{Early}}\right|
          \left|\mathcal{P}_x\right|\Bigr)^{-1} \\
&\quad \times
\sum_{\ell\in\mathcal{L}_{\mathrm{Early}}}
\sum_{t\in\mathcal{P}_x}
\mathrm{TopK}^{(\ell,m)}_{x,t}.
\end{aligned}
\]

\subsubsection{Combining C1 and Paired Differences}
\noindent
\textbf{Combining C1.}
We treat $C1$ as a combined metric consisting of two components,
$\mathrm{C1\_Hoyer}$ and $\mathrm{C1\_TopK}$:
\begin{itemize}
  \item In Sections~4.1--5.1 of the main paper, almost all figures and conclusions
  are based on $\mathrm{C1\_Hoyer}$.
  \item $\mathrm{C1\_TopK}$ and its relationship with Hoyer are reported in
  supplementary figures/tables in Appendix~C, mainly to verify that
  ``densification'' is not an artifact caused by extreme amplification of only a
  few dimensions.
\end{itemize}

\noindent
\textbf{Paired differences.}
At the instance level, we focus on paired differences between a given prefix
condition and the control. For an instance $x$, the Hoyer difference between
\texttt{tag\_correct} and \texttt{control} is defined (consistent with
Section~4.4) as:
\[
\Delta \mathrm{Hoyer}_x
= \mathrm{C1\_Hoyer}\bigl(x,\texttt{tag\_correct}\bigr)
- \mathrm{C1\_Hoyer}\bigl(x,\texttt{control}\bigr).
\]
In implementation, $\Delta \mathrm{C1}$ corresponds to the combination of
$\Delta \mathrm{Hoyer}$ and $\Delta \mathrm{TopK}$ (we report one or both
depending on the figure/table). In the main paper, ``densification'' typically
refers to the case $\Delta \mathrm{Hoyer}_x < 0$.

\subsection{Auxiliary Metrics for Activation-Distribution Shape}
To characterize activation distributions in a more fine-grained manner, we
additionally compute the following metrics for each token-level vector
$\mathbf{h}\in\mathbb{R}^d$:
\begin{itemize}
  \item \textbf{Gini coefficient:} computed from the sequence
  $\bigl(|h_i|\bigr)_{i=1}^d$ using the standard definition, measuring inequality
  in energy allocation across dimensions;
  \item \textbf{Kurtosis:} quantifies whether the distribution exhibits sharp
  peaks and heavy tails;
  \item \textbf{Positive ratio:}
  \[
  \mathrm{PosRatio}(\mathbf{h})
  = \frac{1}{d}\sum_{i=1}^{d} \mathbf{1}\!\left[h_i > 0\right],
  \]
  capturing the sign pattern of activations.
\end{itemize}
These metrics can likewise be averaged over layer segments and used to form
paired differences, but they are only used for robustness checks in
Appendix~C and are not treated as primary evidence for the main-paper
conclusions.

\subsection{Cross-Layer Energy Concentration (Layer-Energy Gini)}

To examine whether meta prompts alter the allocation of ``computational load''
across layers, we define a cross-layer energy Gini metric as follows.
At a key position in the target sequence (e.g., prompt-last or first-gen), we
compute the $\ell$-th layer's $\ell_2$ energy:
\[
E_{\ell}(x,m)
= \left\|\mathbf{h}^{(\ell)}_{x,t^{\star},m}\right\|_2^2,
\qquad \ell = 1,\dots,L,
\]
where $t^{\star}$ denotes the token position of interest.

We then normalize the energies across layers:
\[
\tilde{E}_{\ell}(x,m)
= \frac{E_{\ell}(x,m)}{\sum_{\ell'=1}^{L} E_{\ell'}(x,m)}.
\]

Treating $\{\tilde{E}_{\ell}(x,m)\}_{\ell=1}^{L}$ as a one-dimensional discrete
distribution, we compute its Gini coefficient, denoted as
$\mathrm{Gini}_{\mathrm{layer}}(x,m)$.
Larger values indicate that energy is highly concentrated in a small subset of
layers, whereas smaller values indicate a more even distribution across layers.
This metric is used for supplementary analyses in Appendix~C but is not directly
referenced in the main paper.

\subsection{C2: Domain-Keyword Attention}
Section~4.2 of the main paper defines C2 as the attention share to
domain keywords in the last layer, estimated from two viewpoints:
\textsc{Prompt-last} and \textsc{First-gen}. This section presents the formal
definition.

Let $\mathcal{P}_x$ denote the set of prompt-token positions for an instance
$x$, and let $\mathcal{K}_x \subseteq \mathcal{P}_x$ denote the subset of
positions matched by domain keywords. Under prefix condition $m$, we denote the
attention vector at the last layer ($\ell = L$) for a query position $q$ by
$\mathbf{a}^{(L,m)}_{x,q} \in \mathbb{R}^{|\mathcal{P}_x|}$, where the element
$a^{(L,m)}_{x,q,t}$ is the attention weight assigned to position $t$.

We first renormalize attention over the set of visible tokens
$\mathcal{V}_x \subseteq \mathcal{P}_x$ such that
\[
\sum_{t \in \mathcal{V}_x} a^{(L,m)}_{x,q,t} = 1 .
\]
The domain-keyword attention share at query position $q$ is then defined as
\[
\mathrm{C2}(x,m,q) = \sum_{t \in \mathcal{K}_x} a^{(L,m)}_{x,q,t} .
\]

In the main paper, we consider two query viewpoints:
\begin{itemize}
  \item \textsc{Prompt-last}: set $q = q_{\mathrm{promptlast}}$ as the last
        input-token position, corresponding to the state after ``reading the
        question'';
  \item \textsc{First-gen}: set $q = q_{\mathrm{firstgen}}$ as the first
        generated-token position, corresponding to the state when ``starting to
        answer''.
\end{itemize}

For each viewpoint, we average over instances:
\[
\mathrm{C2}^{\mathrm{PromptLast}}(m)
= \frac{1}{|X|} \sum_{x \in X} \mathrm{C2}(x,m,q_{\mathrm{promptlast}}),
\]
and analogously for the \textsc{First-gen} variant.

At the instance level, paired differences follow the $\Delta$ notation in
Section~4.4. For example,
\[
\Delta \mathrm{C2}^{\mathrm{PromptLast}}_x
= \mathrm{C2}(x,m,q_{\mathrm{promptlast}})
- \mathrm{C2}(x,\mathrm{control},q_{\mathrm{promptlast}}).
\]
A positive $\Delta \mathrm{C2}$ indicates that prefix $m$ increases the
attention share to domain keywords relative to the control.

\paragraph{Note.}
In some figures/tables in Appendix~C, we further extend C2 to
Early/Middle/Late segments by averaging attention across multiple layers.
However, the core results shown in Sections~4.2--5.2 of the main paper are
based on the last-layer definition above.

If an instance has no keyword matches in its domain
($\mathcal{K}_x = \emptyset$), we define $\mathrm{C2}(x,m,q)=0$ and track the
proportion of such ``missing-keyword'' instances separately in the summary
tables.

\subsection{C3: Output Stability}
Section~4.3 of the main paper defines C3 as a pair of measures:
predictive entropy and semantic variation. This section provides the
formal definitions.

\subsubsection{Predictive Entropy}
For an instance $x$ under a prefix condition $m$, let $p^{(m)}_t(v)$ denote
the predictive distribution over vocabulary items $v$ for the $t$-th generated
token. The token-level predictive entropy is
\[
H_t(x,m) \;=\; -\sum_{v} p^{(m)}_t(v)\,\log p^{(m)}_t(v).
\]
Let $T_x$ be the generated sequence length. The sequence-level average entropy is
\[
\bar{H}(x,m) \;=\; \frac{1}{T_x}\sum_{t=1}^{T_x} H_t(x,m).
\]
We use $\bar{H}(x,m)$ as the first C3 measure: lower values indicate a more
concentrated predictive distribution and hence higher confidence.

At the instance level, we define the paired difference
\[
\Delta\mathrm{Entropy}_x \;=\; \bar{H}(x,m) - \bar{H}(x,\mathrm{control}),
\]
where $m$ can be a prefix condition such as \texttt{tag\_correct} or
\texttt{instr\_expert}. The $\Delta\mathrm{Entropy}$ reported in
Section~5.3 refers to this quantity.

\paragraph{Note.}
In implementation, we compute $p^{(m)}_t(v)$ directly from a single forward-pass
softmax. Alternatively, one may average distributions over multiple samples; the
qualitative conclusions remain essentially the same.

\subsubsection{Semantic Variation}
Semantic variation is computed via repeated sampling and sentence-embedding
similarity. For the same prompt-condition pair $(x,m)$, we generate $K$
independent outputs $y^{(1)}(x,m),\dots,y^{(K)}(x,m)$ and encode them using a
fixed sentence encoder $f(\cdot)$:
\[
\mathbf{e}^{(k)}(x,m) \;=\; f\!\big(y^{(k)}(x,m)\big), \quad k=1,\dots,K.
\]
We then compute all pairwise cosine similarities
\[
\begin{aligned}
s_{k\ell}(x,m)
&= \frac{\mathbf{e}^{(k)}(x,m)\cdot \mathbf{e}^{(\ell)}(x,m)}
{\|\mathbf{e}^{(k)}(x,m)\|\,\|\mathbf{e}^{(\ell)}(x,m)\|} \\[2pt]
&\qquad 1 \le k < \ell \le K .
\end{aligned}
\]

We treat the average pairwise similarity as a proxy for semantic F1 (mF1):
\[
\mathrm{mF1}(x,m)
\;=\;
\frac{2}{K(K-1)}
\sum_{1 \le k < \ell \le K} s_{k\ell}(x,m).
\]
Semantic variation is then defined as
\[
\mathrm{Var}(x,m) \;=\; 1 - \mathrm{mF1}(x,m).
\]
Smaller $\mathrm{Var}(x,m)$ indicates that multiple generations are closer in
semantic space and thus more stable. The paired difference is
\[
\Delta\mathrm{Var}_x
\;=\;
\mathrm{Var}(x,m) - \mathrm{Var}(x,\mathrm{control}).
\]

Following Section~4.3, concrete choices of the encoder, the sampling count $K$,
and other hyperparameters are deferred to Appendix~C. In this paper,
$\mathrm{C3}=\{\mathrm{Entropy},\,\mathrm{Var}\}$, and we analyze
$\Delta\mathrm{Entropy}$ and $\Delta\mathrm{Var}$ separately.

\subsection{Commonsense Confidence Margin (Optional Analysis)}
For multiple-choice commonsense tasks, let $\mathcal{C}$ denote the set of
answer options, and let $c_{\mathrm{gold}}\in\mathcal{C}$ be the correct option.
We define the average probability of an option $c$ as
\[
\bar{p}(c;x,m)
\;=\;
\frac{1}{G_x}\sum_{t\in\mathcal{G}_x} p^{(m)}_t(c),
\]
where $p^{(m)}_t(c)$ is the probability assigned to choosing option $c$ at
position $t$ (the exact scoring procedure is described in Appendix~C), and
$\mathcal{G}_x$ denotes the set of positions used for option scoring.

We define the confidence margin as
\[
\Delta_{\mathrm{conf}}(x,m)
\;=\;
\bar{p}(c_{\mathrm{gold}};x,m)
\;-\;
\max_{\substack{c\in\mathcal{C}\\ c\neq c_{\mathrm{gold}}}}
\bar{p}(c;x,m).
\]
Larger $\Delta_{\mathrm{conf}}(x,m)$ indicates a stronger confidence advantage
for the correct option and hence lower uncertainty. In Appendix~C.5, we use this
quantity as an auxiliary metric and analyze it alongside the two primary C3
metrics ($\mathrm{Entropy}$ and $\mathrm{Var}$), without separately elaborating
it in the main text.

With these definitions, all terms appearing in Sections~4.1--4.4 of the main
paper---such as C1/C2/C3, $\Delta$Hoyer, Prompt-last, and First-gen---have
precise counterparts in this appendix, and this section can be directly included
in Appendix~B.

\begin{table*}[t]
\centering
\small
\setlength{\tabcolsep}{6pt}
\begin{tabular}{lccc}
\toprule
\textbf{Model} &
\textbf{$\Delta \mathrm{Hoyer}$ Tag--Ctrl} &
\textbf{$\Delta \mathrm{Hoyer}$ Instr--Ctrl} &
\textbf{$\Delta \mathrm{Hoyer}$ Tag--Instr} \\
\midrule
Llama-3.1-8B-Instruct        & $-0.015 \pm 0.001$ & $-0.016 \pm 0.001$ & $+0.001 \pm 0.000$ \\
Mistral-7B-Instruct-v0.2     & $-0.016 \pm 0.001$ & $-0.017 \pm 0.001$ & $+0.000 \pm 0.000$ \\
Qwen3-8B           & $-0.005 \pm 0.000$ & $-0.003 \pm 0.000$ & $-0.002 \pm 0.000$ \\
\bottomrule
\end{tabular}
\caption{Early segment: mean $\pm$ SEM of $\Delta \mathrm{Hoyer}$ under three contrasts.}
\label{tab:c1_early}
\end{table*}

\begin{table*}[t]
\centering
\small
\setlength{\tabcolsep}{6pt}
\begin{tabular}{lccc}
\toprule
\textbf{Model} &
\textbf{$\Delta \mathrm{Hoyer}$ Tag--Ctrl} &
\textbf{$\Delta \mathrm{Hoyer}$ Instr--Ctrl} &
\textbf{$\Delta \mathrm{Hoyer}$ Tag--Instr} \\
\midrule
Llama-3.1-8B-Instruct        & $-0.014 \pm 0.001$ & $-0.015 \pm 0.001$ & $+0.000 \pm 0.000$ \\
Mistral-7B-Instruct-v0.2     & $-0.012 \pm 0.001$ & $-0.015 \pm 0.001$ & $+0.003 \pm 0.000$ \\
Qwen3-8B          & $-0.007 \pm 0.001$ & $-0.009 \pm 0.001$ & $+0.002 \pm 0.000$ \\
\bottomrule
\end{tabular}
\caption{Middle segment: mean $\pm$ SEM of $\Delta \mathrm{Hoyer}$ under three contrasts.}
\label{tab:c1_middle}
\end{table*}
\FloatBarrier

\section{Supplementary Experimental Results }
\label{app:supp}
This appendix summarizes detailed statistical results based on real experimental data (from routereval\_rq1\_multimodel, rq2\_multimodel, and rq3\_multimodel).

\begin{table*}[t]
\centering
\small
\setlength{\tabcolsep}{6pt}
\begin{tabular}{lccc}
\toprule
\textbf{Model} &
\textbf{$\Delta \mathrm{Hoyer}$ Tag--Ctrl} &
\textbf{$\Delta \mathrm{Hoyer}$ Instr--Ctrl} &
\textbf{$\Delta \mathrm{Hoyer}$ Tag--Instr} \\
\midrule
Llama-3.1-8B-Instruct                 & $-0.008 \pm 0.000$ & $-0.012 \pm 0.000$ & $+0.004 \pm 0.000$ \\
Mistral-7B-Instruct-v0.2              & $-0.002 \pm 0.000$ & $-0.015 \pm 0.000$ & $+0.013 \pm 0.000$ \\
Qwen3-8B                    & $-0.001 \pm 0.000$ & $-0.007 \pm 0.001$ & $+0.006 \pm 0.000$ \\
\bottomrule
\end{tabular}
\caption{Global segment: mean $\pm$ SEM of instance-level paired differences in $\Delta \mathrm{Hoyer}$ under three contrasts.}
\label{tab:c1_global}
\end{table*}

\subsection{Densification Results by Model and Layer Segment (RQ1)}
Tables~\ref{tab:c1_early}--\ref{tab:c1_global} report changes in Hoyer sparsity
($\Delta \mathrm{Hoyer}$) for each model across layer segments.
Values are reported as mean $\pm$ standard error of the mean (SEM).
Negative values indicate densification (lower sparsity).

\paragraph{Interpretation note.}
For the contrast Tag--Instr, a \emph{positive} value means
$\mathrm{Hoyer}(\text{Tag}) > \mathrm{Hoyer}(\text{Instr})$,
i.e., expert instructions induce stronger densification than tags.

\subsection{Domain-Keyword Attention (RQ2)}

\subsubsection{RQ2 Statistical Details}
\label{app:rq2_stats}
For RQ2, we compute instance-level paired differences in the domain-keyword
attention share,
$\Delta \mathrm{Attn}=\mathrm{Attn}(\text{condition})-\mathrm{Attn}(\text{control})$,
under two query viewpoints (Prompt-last and First-gen).
Tables~\ref{tab:rq2_attn_promptlast}--\ref{tab:rq2_attn_firstgen} summarize these
effects as mean $\pm$ standard error of the mean (SEM).
When the text describes an effect as statistically significant, it is based on
paired $t$-tests of $\Delta \mathrm{Attn}$ against zero with Benjamini--Hochberg
FDR control within each viewpoint/table family (test statistics and adjusted
$p$-values are omitted from the tables for space).

To relate attention changes to output stability, we compute instance-level
Pearson correlations between $\Delta \mathrm{Attn}$ and $\Delta \mathrm{C3}$
metrics; Table~\ref{tab:rq2_attn_entropy_corr} reports Pearson's $r$ with
two-sided $p$-values.

\subsubsection{$\Delta$Attn from the Prompt-last View}
Table~\ref{tab:rq2_attn_promptlast} reports paired differences in the domain-keyword attention share under the Prompt-last view
($\Delta \mathrm{Attn} = \text{condition} - \text{control}$).
Negative values indicate a reduced attention share to domain keywords under the condition, whereas positive values indicate an increased share.
Values are reported as mean $\pm$ standard error of the mean (SEM).

\begin{table*}[t]
  \centering
  \small
  \setlength{\tabcolsep}{6pt}
  \begin{tabular}{lcc}
    \toprule
    \textbf{Model} &
    \textbf{$\Delta \mathrm{Attn}$ Tag--Ctrl (Mean $\pm$ SEM)} &
    \textbf{$\Delta \mathrm{Attn}$ Instr--Ctrl (Mean $\pm$ SEM)} \\
    \midrule
    Llama-3.1-8B-Instruct    & $-0.0389 \pm 0.0031$ & $-0.0371 \pm 0.0030$ \\
    Mistral-7B-Instruct-v0.2 & $+0.0154 \pm 0.0005$ & $+0.0164 \pm 0.0005$ \\
    Qwen3-8B                 & $-0.0409 \pm 0.0032$ & $-0.0396 \pm 0.0031$ \\
    \bottomrule
  \end{tabular}
  \caption{Mean $\pm$ SEM of $\Delta \mathrm{Attn}$ under Tag--Ctrl and Instr--Ctrl across three models (Prompt-last view).}
  \label{tab:rq2_attn_promptlast}
\end{table*}

\subsubsection{$\Delta$Attn from the First-gen View}
Table~\ref{tab:rq2_attn_firstgen} reports paired differences in the domain-keyword attention share under the First-gen view
(i.e., at the first generated token), defined in the same way as Table~\ref{tab:rq2_attn_promptlast}.
\begin{table*}[t]
  \centering
  \small
  \setlength{\tabcolsep}{6pt}
  \begin{tabular}{lcc}
    \toprule
    \textbf{Model} &
    \textbf{$\Delta \mathrm{Attn}$ Tag--Ctrl (Mean $\pm$ SEM)} &
    \textbf{$\Delta \mathrm{Attn}$ Instr--Ctrl (Mean $\pm$ SEM)} \\
    \midrule
    Llama-3.1-8B-Instruct    & $-0.0592 \pm 0.0045$ & $-0.0599 \pm 0.0046$ \\
    Mistral-7B-Instruct-v0.2 & $+0.0158 \pm 0.0004$ & $+0.0175 \pm 0.0005$ \\
    Qwen3-8B                 & $-0.0493 \pm 0.0039$ & $-0.0480 \pm 0.0038$ \\
    \bottomrule
  \end{tabular}
  \caption{Mean $\pm$ SEM of $\Delta \mathrm{Attn}$ under Tag--Ctrl and Instr--Ctrl across three models (First-gen view). Negative values indicate reduced domain-keyword attention share relative to control.}
  \label{tab:rq2_attn_firstgen}
\end{table*}

We observe that:
\begin{itemize}
  \item For Llama and Qwen, $\Delta \mathrm{Attn}$ is significantly negative under both views (reduced reliance on keywords), roughly consistent with ``cognitive offloading.''
  \item For Mistral, $\Delta \mathrm{Attn}$ is significantly positive under both views (increased reliance on keywords; ``attention reinforcement''), with a slightly larger magnitude under Instr than Tag.
\end{itemize}

\subsubsection{Correlation Between $\Delta$Attn and Delta Entropy}
Table~\ref{tab:rq2_attn_entropy_corr} reports the instance-level Pearson correlation coefficient between First-gen $\Delta \mathrm{Attn}$
and $\Delta \mathrm{Entropy}$ under the \texttt{instr\_expert} vs.\ \texttt{control} contrast.
Here, $\Delta \mathrm{Attn}$ uses
\texttt{attn\_\allowbreak domain\_\allowbreak share\_\allowbreak firstgen\_\allowbreak
delta\_\allowbreak instrexpert\_\allowbreak vs\_\allowbreak control}, and
$\Delta \mathrm{Entropy}$ uses
\texttt{gen\_\allowbreak mean\_\allowbreak entropy\_\allowbreak delta\_\allowbreak
instrexpert\_\allowbreak vs\_\allowbreak control}.

\begin{table*}[t]
  \centering
  \small
  \setlength{\tabcolsep}{8pt}
  \begin{tabular}{lcc}
    \toprule
    \textbf{Model} &
    \textbf{$r(\Delta \mathrm{Attn},\, \Delta \mathrm{Entropy})$} &
    \textbf{$p$-value} \\
    \midrule
    Llama-3.1-8B-Instruct    & $0.117$  & $1.3 \times 10^{-10}$ \\
    Mistral-7B-Instruct-v0.2 & $-0.072$ & $7.8 \times 10^{-5}$  \\
    Qwen3-8B                 & $0.229$  & $5.4 \times 10^{-37}$ \\
    \bottomrule
  \end{tabular}
  \caption{Pearson correlation between $\Delta \mathrm{Attn}$ and $\Delta \mathrm{Entropy}$ (instance-level), with corresponding $p$-values.}
  \label{tab:rq2_attn_entropy_corr}
\end{table*}

Brief interpretation:
\begin{itemize}
  \item Qwen shows the highest correlation ($r \approx 0.23$), suggesting a relatively stronger linkage between keyword-attention changes and entropy changes in this model.
  \item Llama exhibits a moderate-to-weak positive correlation ($r \approx 0.12$).
  \item Mistral shows a significant negative correlation, consistent with its ``attention reinforcement'' pattern accompanied by limited gains in stability.
\end{itemize}

\subsection{Correlations Between C1 and C3 (RQ3)}
\paragraph{Notation.}
{\raggedright
\begin{itemize}
  \item \textbf{C1}: \texttt{c1\_prompt\_hoyer\_mean} (paired difference in Hoyer sparsity over the prompt segment).
  \item \textbf{C3-Entropy}: paired difference of \texttt{gen\_mean\_entropy}.
  \item \textbf{C3-SemVar}: paired difference of \texttt{semantic\_variation\_1mF1}.
  \item \textbf{Instr vs.\ Ctrl}: \texttt{instrexpert\_vs\_control}.
  \item \textbf{Tag vs.\ Ctrl}: \texttt{tagcorrect\_vs\_control}.
  \item \textbf{Tag vs.\ Instr}: \texttt{tagcorrect\_vs\_instrexpert}.
  \item \textbf{Significance markers} appended to correlation coefficients:
  $^{*}p<0.05$, $^{**}p<0.01$, $^{***}p<0.001$, and ns for not significant.
\end{itemize}
}

\subsubsection{RQ3 Statistical Details}
\label{app:rq3_stats}
We report Pearson correlations between paired differences
$\Delta \mathrm{C1}$ (prompt Hoyer) and $\Delta \mathrm{C3}$ (entropy or semantic variation)
for each model and prefix contrast. Two-sided $p$-values are computed per correlation;
multiple comparisons are controlled with Benjamini--Hochberg correction within each
table family. Full results are in Tables~\ref{tab:rq3_corr_entropy}--~\ref{tab:rq3_corr_semvar}.

\subsubsection{Correlation Between C1 and C3-Entropy}
Table~\ref{tab:rq3_corr_entropy} reports Pearson correlation coefficients between
C1 ($\Delta \mathrm{Hoyer}$) and C3-Entropy ($\Delta \mathrm{Entropy}$).

\begin{table*}[t]
  \centering
  \small
  \setlength{\tabcolsep}{8pt}
  \begin{tabular}{lccc}
    \toprule
    \textbf{Model} &
    \textbf{$r$ (Instr vs.\ Ctrl)} &
    \textbf{$r$ (Tag vs.\ Ctrl)} &
    \textbf{$r$ (Tag vs.\ Instr)} \\
    \midrule
    Llama-3.1-8B-Instruct    & $0.141^{***}$ & $0.013^{\mathrm{ns}}$ & $0.105^{***}$ \\
    Mistral-7B-Instruct-v0.2 & $0.191^{***}$ & $0.043^{*}$           & $0.150^{***}$ \\
    Qwen3-8B                 & $0.309^{***}$ & $0.204^{***}$         & $0.139^{***}$ \\
    \bottomrule
  \end{tabular}
  \caption{Pearson correlations between C1 ($\Delta \mathrm{Hoyer}$) and C3-Entropy
  ($\Delta \mathrm{Entropy}$) under three contrasts. Superscripts indicate significance:
  $^{***}p<0.001$, $^{**}p<0.01$, $^{*}p<0.05$, and ns not significant.}
  \label{tab:rq3_corr_entropy}
\end{table*}

\noindent
\textit{Rounding.} Values correspond to \texttt{pearson\_r} in the CSV; e.g., for Qwen3-8B under
Instr vs.\ Ctrl, $r=0.308859$ rounds to $0.309$.

\subsubsection{Correlation Between C1 and C3-SemVar}
Table~\ref{tab:rq3_corr_semvar} reports Pearson correlation coefficients between
C1 ($\Delta \mathrm{Hoyer}$) and C3-SemVar ($\Delta \mathrm{Var}$, i.e., \texttt{semantic\_variation\_1mF1}).


\begin{table}[t]
  \centering
  \scriptsize
  \setlength{\tabcolsep}{3pt} 
  \begin{tabular}{@{}lccc@{}}
    \toprule
    \textbf{Model} &
    \textbf{Instr--Ctrl} &
    \textbf{Tag--Ctrl} &
    \textbf{Tag--Instr} \\
    \midrule
    Llama-3.1-8B-Instruct    & $0.067^{***}$         & $-0.027^{\mathrm{ns}}$ & $0.037^{*}$ \\
    Mistral-7B-Instruct-v0.2 & $0.142^{***}$         & $0.128^{***}$          & $0.177^{***}$ \\
    Qwen3-8B                 & $0.023^{\mathrm{ns}}$ & $0.072^{***}$          & $0.017^{\mathrm{ns}}$ \\
    \bottomrule
  \end{tabular}
  \caption{Pearson $r$ between C1 ($\Delta \mathrm{Hoyer}$) and C3-SemVar ($\Delta \mathrm{Var}$).
  Superscripts: $^{***}p<0.001$, $^{**}p<0.01$, $^{*}p<0.05$, $\mathrm{ns}$ not significant.}
  \label{tab:rq3_corr_semvar}
\end{table}

\section{Reproducibility and Resource Consumption}
\subsection{Hardware and Runtime}
\begin{itemize} 
\item All experiments are conducted on multiple high-memory GPUs (8×A100). We disable gradient computation and only perform forward inference with activation/attention caching;
\item For a single model, running MFV and extracting the metrics on the RouterEval subset used in this paper takes on the order of several hours, mainly depending on the model size and the number of stochastic generations (K);
\item The analysis scripts (C1–C3 computation and statistical tests) are primarily bottlenecked by I/O and aggregation, and typically take less time than forward inference.
\end{itemize}

\subsection{Randomness and Reproducibility}
\begin{itemize}
\item All stochastic components (instance shuffling, incorrect-tag sampling, and decoding-time sampling) use fixed random seeds;
\item For stochastic metrics such as semantic variation, we report the variance across repeated runs in the appendix;
\item When supported by the framework, we enable deterministic options as much as possible to reduce the impact of numerical nondeterminism on the results.
\end{itemize}

\subsection{Code and Data Release Plan}
During the anonymous review phase, we do not disclose the repository URL. If the paper is accepted, we plan to:
\begin{itemize}
\item Release the RouterEval preprocessing scripts and the routing-style meta-prompt templates;
\item Release the code for C1–C3 computation and statistical analysis, including MFV scripts for multi-GPU parallelization;
\item Provide configuration files and command-line examples to facilitate reproducing experiments and figures;
\item Provide, within licensing constraints, the indices or hashes of the RouterEval subset used, enabling other researchers to compare results on the same set of instances.
\end{itemize}

\section{Extended Ethics Discussion}
When deploying a similar analysis framework in real-world systems, the following ethical and safety considerations should be taken into account:
\begin{enumerate}
\item Potential manipulation risks
 A finer-grained understanding of how routing signals relate to internal representations could be misused to more precisely steer model behavior or to circumvent safety mechanisms. Our work is intended for diagnosis and robustness analysis; we do not endorse any application of RIDE or related techniques for deceptive, surveillance-oriented, or otherwise harmful purposes.
\item Bias and unfair routing
 If ``densification--stability'' signals are directly used as routing cues or as proxies for uncertainty, they may exhibit systematic differences across languages, demographic groups, or topics, thereby amplifying biases present in the training data. Future work should evaluate the fairness properties of RIDE metrics on multilingual and multi-group datasets.
\item Privacy and data usage
 This paper uses only public benchmarks and open-source models, without involving real user data. If future work extends the framework to real routing logs or production data, strict anonymization should be applied, and relevant privacy regulations and platform policies must be followed.
\item Cautious claims about ``interpretability''
 We emphasize that RIDE provides a statistical lens for interpretation rather than a strict causal characterization of underlying mechanisms. In external communication or system documentation, it should not be presented as a ``fully interpretable'' solution; instead, its role as an analysis tool and auxiliary signal should be stated explicitly.
\end{enumerate}

\end{document}